\documentclass[11pt]{article}
\usepackage{emnlp2018}
\usepackage{times}
\usepackage{latexsym}
\usepackage{url}
\usepackage{examples}
\usepackage{multirow}
\usepackage{graphicx}
\usepackage{amsmath}
\usepackage[para,online,flushleft]{threeparttable}
\usepackage{tablefootnote}

\aclfinalcopy 

\title{
A Deterministic Algorithm for Bridging Anaphora Resolution  
}

\author{
Yufang Hou \\
  IBM Research Ireland\\ {\tt yhou@ie.ibm.com}\\
   }

\date{}
\begin{document}
\maketitle

\begin{abstract}
Previous work on bridging anaphora resolution \cite{poesio04d,houyufang13a} use syntactic preposition patterns to calculate word relatedness. 
However, such patterns only consider NPs' head nouns and hence do not fully capture the semantics of NPs.  
Recently, \newcite{houyufang18} created word embeddings (\emph{embeddings\_PP}) to capture associative similarity (i.e., relatedness) between nouns by exploring the syntactic structure of noun phrases.
But \emph{embeddings\_PP} only contains word representations for nouns.  
In this paper, we create new word vectors by combining \emph{embeddings\_PP} with \emph{GloVe}. 
This new word embeddings (\emph{embeddings\_bridging}) are a more general lexical knowledge resource for bridging and allow us to represent the meaning of an NP beyond its head easily. 
We therefore develop a deterministic approach for bridging anaphora resolution, which represents 
the semantics of an NP based on its head noun and modifications. 
We show that this simple approach achieves the competitive results compared to the best system in \newcite{houyufang13a} which explores Markov Logic Networks to model the problem. 
Additionally, we further improve the results for bridging anaphora resolution reported in \newcite{houyufang18} by combining our simple deterministic approach with \newcite{houyufang13a}'s best system \emph{MLN II}.

\end{abstract}

\section{Introduction}
\label{sec:intro}


Anaphora plays a major role in discourse comprehension and 
accounts for the coherence of a text.
In contrast to  
\emph{identity anaphora} which indicates that a noun phrase refers back to 
the same entity introduced by previous descriptions in the discourse,
\emph{bridging anaphora} or \emph{associative anaphora} links anaphors and antecedents via lexico-semantic,
frame or encyclopedic relations.
\emph{Bridging resolution}\ is the task to recognize bridging anaphors (e.g., \textbf{distribution arrangements} in Example \ref{ex:bridging2}\footnote{All examples, if not specified
  otherwise, are from ISNotes \cite{markert12}.  Bridging
  anaphors are typed in boldface, antecedents in italics
  throughout this paper.})
and find links to their antecedents (e.g., \emph{dialysis products} in Example \ref{ex:bridging2}). 

\begin{examples}
\item \label{ex:bridging2} While the discussions between Delmed and National Medical Care have been discontinued, Delmed will continue to supply \emph{dialysis products} through National Medical after their exclusive agreement ends in March 1990, Delmed said.\\
In addition, Delmed is exploring \textbf{distribution arrangements} with Fresenius USA, Delmed said.
\end{examples}

Most previous empirical research on bridging \cite{poesio98,poesio04d,markert03,lassalle11,houyufang13a}
focus on \emph{bridging anaphora resolution}, a subtask of bridging resolution that aims to choose the antecedents for bridging anaphors.
For this substask, most previous work \cite{poesio04d,lassalle11,houyufang13a}
calculate semantic relatedness between an anaphor and its antecedent based on word co-occurrence counts using certain syntactic patterns. However, such patterns only consider
head noun knowledge and hence are not sufficient for bridging relations which require the semantics of modification. In Example \ref{ex:bridging2}, in order to find the antecedent (\emph{dialysis products})
for the bridging anaphor ``\textbf{distribution arrangements}'',  we have to understand the semantics of the modification ``\textbf{distribution}''.

Over the past few years, 
word embeddings gained a lot popularity in the NLP community. 
State-of-the-art word vectors such as word2vec skip-gram \cite{mikolov13} and GloVe \cite{pennington14} have been shown to perform well across a variety of NLP tasks,
including textual entailment \cite{tim16}, reading comprehension \cite{chendanqi16}
and coreference resolution \cite{leekenton17}. 

Recently, \newcite{houyufang18} found that these vanilla word embeddings capture both ``genuine'' similarity and relatedness, and hence they are not suitable for bridging anaphora resolution which 
requires lexical association knowledge instead of semantic similarity information between synonyms
or hypernyms. \newcite{houyufang18} created word embeddings for bridging (\emph{embeddings\_PP}) by exploring 
the syntactic structure of noun phrases (NPs) to derive contexts for nouns in the GloVe model. 

However, \emph{embeddings\_PP} only contains the word representations for nouns. 
In this paper, we improve \emph{embeddings\_PP} by combining it with \emph{GloVe}.
The resulting word embeddings (\emph{embeddings\_bridging}) are a more general lexical knowledge resource for bridging anaphora resolution. 
Compared to \emph{embeddings\_PP},
the coverage of lexicon in \emph{embeddings\_bridging} is much larger. Also the word representations for nouns without the suffix ``\_PP'' are more
accurate because they are trained on many more instances in the vanilla \emph{GloVe}. 
Based on this general vector space, we develop a deterministic algorithm to select antecedents for bridging anaphors. 
Our approach combines the semantics of an NP's head with the semantics of its modifications by vector average using \emph{embeddings\_bridging}. 
We show that this simple, efficient method achieves the competitive results on ISNotes for the task of bridging anaphora resolution compared to the best system in \newcite{houyufang13a} which explores Markov Logic Networks to model the problem.

The main contributions of our work are: (1) a general word representation resource\footnote{\emph{embeddings\_bridging} can be downloaded from \url{https://doi.org/10.5281/zenodo.1403164}} for bridging; and (2) a simple 
yet competitive deterministic approach for bridging anaphora resolution which models the meaning
of an NP based on its head noun and modifications. 


\begin{table*}[htb]
 \begin{center}
\begin{tabular}{l|l}
\hline
\textbf{Noun Phrases}& \textbf{Extracted Noun Pairs}\\ \hline
\emph{\textbf{travelers} in the train station} &\emph{travelers\_PP -- station} \\
\emph{\textbf{travelers} from the airport} &\emph{travelers\_PP -- airport} \\
\emph{\textbf{hotels} for travelers}& \emph{hotels\_PP -- travelers}\\
\emph{the \textbf{destination} for travelers} & \emph{destination\_PP -- travelers}\\
\emph{the company's new appointed \textbf{chairman}} & \emph{chairman\_PP -- company}\\
\hline
\end{tabular}
\end{center}
\caption{\label{tab:embeddingppexample} Examples of noun phrases as well as the extracted noun pairs in \emph{embeddings\_PP}. Bold indicates the head noun of an NP.}
\end{table*}

\section{Related Work}
\label{sec:relatedwork}

\paragraph{Lexical/world knowledge for bridging:} 
\newcite{houyufang13b} explored various lexico-semantic features for bridging anaphora recognition. 
\newcite{houyufang16} proposed an attention-based LSTM model with pre-trained word embeddings
for information status classification and reported moderate results for bridging recognition. 
Previous work on bridging anaphora resolution \cite{poesio04d,lassalle11,houyufang13a} explored word co-occurrence counts in certain syntactic preposition patterns to calculate word relatedness.
For instance, the big hit counts of the query
``\emph{the door of the house}'' in large corpora could indicate that \emph{door} and \emph{house} stand in a part-of relation. 
These patterns encode associative relations between nouns which cover a variety of bridging relations. 
Unlike previous work which only consider a small number of prepositions per anaphor, 
the PP context model \cite{houyufang18} uses all prepositions for all nouns in big corpora. It also includes the possessive structure of NPs. 
In this paper, we further improve \newcite{houyufang18}'s \emph{embeddings\_PP} by combining it with the vanilla \emph{GloVe}. 
The resulting word embeddings (\emph{embeddings\_bridging}) are a more general lexical knowledge resource for bridging resolution. 
In addition, it enables efficient computation of word association strength through low-dimensional matrix operations.

\paragraph{Bridging anaphora resolution:} 
regarding the methods to select antecedents for bridging anaphors, \newcite{poesio04d} applied a pairwise model combining lexical semantic features as well as salience features
to perform mereological bridging resolution in the GNOME corpus\footnote{The GNOME corpus is not publicly available.}. To address the data sparseness
problem (e.g., some part-of relations are not covered by WordNet), they used the Web
to estimate the part-of relations expressed by certain syntactic constructions. 
Based on the method proposed by \newcite{poesio04d},  
\newcite{lassalle11} developed a system that resolves mereological bridging
anaphors in French. The system was enriched with meronymic information extracted from raw texts. Such information
was extracted in a bootstrapping fashion by iteratively collecting 
meronymic pairs and the corresponding syntactic patterns.
\newcite{lassalle11} evaluated their system on mereological bridging anaphors annotated in the DEDE corpus
and reported an accuracy of 23\%.

\newcite{markert12} released a corpus called ISNotes which contains unrestricted bridging annotations. 
Based on this corpus, \newcite{houyufang13a} proposed a joint inference framework for bridging anaphora resolution using Markov logic networks \cite{domingos09}. 
The framework resolves all bridging anaphors in one document together by modeling that semantically related
anaphors are likely to share the same antecedent.

ISNotes is a challenging corpus for bridging. First, bridging
anaphors are not limited to definite NPs as in previous work \cite{poesio97b,poesio04d,lassalle11}. 
Also in ISNotes, the semantic relations between anaphor and antecedent are not restricted to meronymic relations.
We therefore choose ISNotes to evaluate our algorithm for bridging anaphora resolution. Our approach
is deterministic and simple, but achieves the competitive results compared to the advanced machine learning-based approach \cite{houyufang13a}.
We also improve the result reported in \newcite{houyufang18} on the same corpus by combining our deterministic approach with the best system from \newcite{houyufang13a}.

Just recently, two new corpora \cite{roesiger18b, poesio18} with bridging annotations have become available and we notice that  
the definitions of bridging in these corpora are different from the bridging definition in ISNotes. 
We apply our algorithm with small adaptations to select antecedents for bridging anaphors on these corpora. The moderate results demonstrate that \emph{embeddings\_bridging}
is a general word representation resource for bridging.

\section{Word Representations for Bridging}
\label{sec:method}

%

\subsection{Word Embeddings Based on PP Contexts (\emph{embeddings\_PP})}
\label{subsec:embiddingspp}
We briefly describe \newcite{houyufang18}'s \emph{embeddings\_PP} in this section. 
\emph{embeddings\_PP} released by \newcite{houyufang18} contains 100-dimensional vectors for 276k nouns.
It is trained over 197 million noun pairs extracted from the automatically parsed
Gigaword corpus \cite{gigaword5.0data,napoles12}. The author generates these noun pairs by exploring the syntactic prepositional and possessive structures of noun phrases. 
These two structures encode a variety of bridging relations between anaphors and their antecedents. For instance, the prepositional structure in ``\emph{the door of the house}'' indicates the part-of relation between ``\emph{door}'' 
and ``\emph{house}''.
More specifically, for NPs containing the prepositional structure (e.g.,  \textbf{X} \emph{preposition} \textbf{Y}) or the possessive structure (e.g., \textbf{Y} \emph{'s} \textbf{X}), 
the author extracts the noun pair ``X\_PP--Y''. Note that the head of the NP is always on the left and the noun modifier is always on the right. In addition, the suffix ``\_PP'' 
is added for the nouns on the left. 
Table \ref{tab:embeddingppexample} shows a few examples of noun phrases together with the extracted noun pairs.

\newcite{houyufang18} showed that the suffix ``\_PP'' plays an important role for the model to learn the asymmetric relations between the head nouns and their noun modifiers from the extracted noun pairs. 
For instance, among the top five nearest neighbors in \emph{embeddings\_PP}, ``president\_PP'' is mostly related to countries or organizations (e.g., ``federation'',
``republic'', or ``USA''), while ``president'' is mostly related to words which have the same semantic type as ``president'' (e.g., ``minister'', ``mayor'', or ``governor''). 

\begin{table*}[htb]
 \begin{center}
\begin{tabular}{l|c|c}
\hline
& \textbf{\emph{embeddings\_bridging}} & \textbf{\emph{embeddings\_PP}} \\ \hline
dimension size&200&100 \\
vocabulary size&532,768&276,326 \\
word type & all& nouns\\
\hline
\end{tabular}
\end{center}
\caption{\label{tab:embeddingcompare}  Comparison between \emph{embeddings\_bridging} and \emph{embeddings\_PP}}
\end{table*}

\begin{table*}[thb]
\begin{center}
\begin{tabular}{@{}l|l|l|c|c|c@{}}
\hline
\textbf{Category}&\textbf{Relation}&\textbf{prototypical word}&\textbf{\emph{embeddings}}&\textbf{\emph{embeddings}}&\textbf{GloVe}\\ 
&&\textbf{pair example}&\textbf{\emph{\_bridging}}&\textbf{\emph{\_PP}}&\\ 
\hline
\emph{PART--WHOLE}&\emph{Object: Component}&\{face: nose\}&\textbf{0.43}&0.27&0.40\\
\emph{PART--WHOLE}&\emph{Event: Feature}&\{wedding: bride\}&\textbf{0.46}&0.29&0.15\\
\emph{PART--WHOLE}&\emph{Creature: Possession}&\{author: copyright\}&\textbf{0.23}&0.14&0.21\\
\emph{PART--WHOLE}&\emph{Activity: Stage}&\{buying: shopping\}&\textbf{0.32}&0.30&0.25\\
\hline
\end{tabular}
\end{center}
\caption{\label{tab:relationalsimilarity} Spearman's rank correlation coefficient $\rho$ for typical \emph{PART--WHOLE} bridging relations using \emph{embeddings\_bridging}, \emph{embeddings\_PP} and \emph{GloVe}.}
\end{table*}

\subsection{Word Representations for Bridging (\emph{embeddings\_bridging})}
\label{subsec:embeddingsbridging}
\emph{embeddings\_PP} described in the previous section only contains word representations for nouns. To improve the coverage of lexical information, we create
a general word representation resource \emph{embeddings\_bridging} by merging \emph{embeddings\_PP} with the original GloVe vectors trained 
on Gigaword and Wikipedia datasets. Specifically, given the 100 dimension word embeddings \emph{embeddings\_PP} and \emph{GloVe}, we first create a 100 dimension vector $v_{filler}$
with the value of each dimension as $0.1$\footnote{Theoretically, any 100 dimension random vector with uniform distribution could be used as $v_{filler}$.}. 
Let $v1_{w}$ represent the vector for the word $w$ in \emph{GloVe}, $v2_{w}$ represent the vector for the word $w$ in \emph{embeddings\_PP},  
if a word $w$ appears both in \emph{GloVe} and in \emph{embeddings\_PP}, its vector in  \emph{embeddings\_bridging} is the concatenation of $v1_{w}$ and $v2_{w}$. 
For the word $w_1$ which only appears in \emph{GloVe}, its vector in  \emph{embeddings\_bridging} is the concatenation of $v1_{w_1}$ and $v_{filler}$. 
Finally, for the word $w_2$ which only appears in \emph{embeddings\_PP} (all the words with the suffix ``\emph{\_PP}''), we construct its vector by concatenating $v_{filler}$
and  $v2_{w_2}$. The resulting 200 dimension word embeddings (\emph{embeddings\_bridging}) is a general lexical resource for bridging. Table \ref{tab:embeddingcompare} compares the main 
features between \emph{embeddings\_bridging} and \emph{embeddings\_PP}. In the next section, we will compare 
\emph{embeddings\_bridging} with \emph{embeddings\_PP} and the original \emph{GloVe} on a few typical bridging relations in the task of measuring relational similarity \cite{jurgens12}.
Moreover, in Section \ref{subsec:eval1} and Section \ref{subsec:eval2}, we show  
that using \emph{embeddings\_bridging} yields better results than using \emph{embeddings\_PP} for bridging anaphora resolution.

\subsection{Measuring Relational Similarity on Typical Bridging Relations}
We evaluate our \emph{embeddings\_bridging} quantitatively using a few typical bridging relations from SemEval-2012 Task 2 \cite{jurgens12}.  
The shared task aims to rank word pairs by the degree to which they are prototypical members of a given relation class.
For instance, given the  prototypical word pairs \emph{\{wedding--bride, rodeo--cowboy, banquet--food\}} for the relation \emph{Event:Feature}, we would like to know among the input word
pairs \emph{\{school--students, circus--clown, meal--food, lion--zoo\}}, which one represents the relation best.

SemEval-2012 Task 2 contains 79 relation classes chosen from \newcite{bejar1991}. These relations fall into ten main categories, including \emph{SIMILAR},
\emph{PART--WHOLE}, \emph{CONTRAST} and more. Each relation class is paired with a few prototypical word pairs and a list of around 40 word pairs which are ranked by humans according to their degree of similarity
to the corresponding relation. We choose all typical bridging relations under the \emph{PART--WHOLE} category and evaluate our \emph{embeddings\_bridging} in terms of ranking the list of word pairs for each relation. 
Spearman's rank correlation coefficient $\rho$ is used to evaluate a system  by comparing the system's ranking of the word pairs against the gold standard ranking.

Following \newcite{zhila13}, we calculate the relational similarity between word pairs using cosine similarity. Let $(w_1, w_2)$ and $(w_3, w_4)$ be the two word pairs,  
$v_1$, $v_2$, $v_3$, $v_4$ be the corresponding vectors for these words. We first normalize all word vectors to unit vectors, then the relational similarity between $(w_1, w_2)$ and $(w_3, w_4)$ is calculated as: 
\begin{equation} \label{eq:pairsimilarity}
\frac{(v1-v2) \cdot (v3-v4)}{\parallel v1-v2\parallel \parallel v3-v4\parallel}  
\end{equation}

\begin{table*}[thb]
\begin{center}
\begin{tabular}{l|l|l|l|l}
\hline
\multicolumn{5}{c}{bridging anaphor: \textbf{distribution arrangements}}\\ \hline
\textbf{Ante. Candidates}&\textbf{Head}& \textbf{Head + Modifiers}&\textbf{$dist_{h}$}&\textbf{$dist_{hm}$}\\ 
\hline
\emph{the discussions between Delmed}&\{\emph{discussions}\}&\{\emph{discussions}\}&0.05&-0.10\\
\emph{and National Medical Care}&&&&\\ \hline
\emph{Delmed}&\{\emph{delmed}\}&\{\emph{delmed}\}&---&---\\ \hline
\emph{National Medical Care}&\{\emph{care}\}&\{\emph{care}\}&\textbf{0.08}&0.10\\ \hline
\emph{dialysis products}&\{\emph{products}\}&\{\emph{dialysis, products}\}&0.06&\textbf{0.17}\\ \hline
\emph{National Medical}&\{\emph{medical}\}&\{\emph{medical}\}&0.02&-0.01\\ \hline
\emph{their}&\{\emph{their}\}&\{\emph{their}\}&-0.05&-0.01\\ \hline
\emph{their exclusive agreement}&\{\emph{agreement}\}&\{\emph{exclusive, agreement}\}&0.07&0.03\\
\hline
\end{tabular}
\end{center}
\caption{\label{tab:similarity} The cosine similarities between the bridging anaphor \textbf{distribution arrangements} and its antecedent candidates for Example \ref{ex:bridging2}. 
$dist_{h}$ indicates the cosine similarity between \{\textbf{arrangements\_PP}\} and the candidate head, $dist_{hm}$ the cosine similarity between \{\textbf{distribution\_PP}, \textbf{arrangements\_PP}\} and Head+Modifiers.
``--'' means \emph{Delmed} is not present in \emph{embeddings\_bridging} and therefore we neglect this candidate.}
\end{table*}

For each chosen relation class, we rank the list of word pairs according to their mean relational similarity to the given prototypical word pairs. 
Table \ref{tab:relationalsimilarity} shows the results of Spearman's rank correlation coefficient $\rho$ for each typical bridging relation using \emph{embeddings\_bridging}, \emph{embeddings\_PP}, and \emph{GloVe}, respectively. 
Note that when using \emph{embeddings\_bridging} and \emph{embeddings\_PP}, we add the suffix ``\_PP'' to the potential bridging anaphor for each word pair (e.g., \{wedding: bride\_PP\}). 
As shown in Table \ref{tab:relationalsimilarity}, using \emph{embeddings\_bridging} performs better than both using \emph{embeddings\_PP} and using the vanilla GloVe vectors on these four part-of relation classes. 
This partially indicates 
that \emph{embeddings\_bridging} could capture 
lexical knowledge for bridging relations.

\section{A Deterministic Algorithm for Bridging Anaphora Resolution}
\label{sec:algorithm}
In this section, we describe our deterministic algorithm based on \emph{embeddings\_bridging} for bridging anaphora resolution. 
For each anaphor $a$, we 
construct the list of antecedent candidates $E_a$ using NPs preceding $a$ from the same sentence as well as 
from the previous two sentences. \newcite{houyufang13a} found that globally salient entities are likely to be the antecedents of all anaphors in a text. We approximate this by adding NPs from the first sentence of 
the text 
to $E_a$. This is motivated by the fact that ISNotes is a newswire corpus and globally salient entities are often introduced in the beginning of an article.
We exclude an NP from $E_a$ if it is a bridging anaphor because a bridging anaphor is rarely to be an antecedent for another bridging anaphor.
We also exclude NPs whose semantic types are ``time'' from $E_a$ if $a$ is not a time expression. This is because time expressions are related to a lot of words in the corpus
in which we learned \emph{embeddings\_bridging} from. Therefore we only keep them as the antecedent candidates for bridging anaphors whose semantic types are ``time'' (see Example \ref{ex:bridging3}). 

\begin{examples}
\item \label{ex:bridging3} As a presidential candidate in \emph{1980}, George Bush forthrightly expressed his position on abortion in an interview with Rolling Stone magazine published \textbf{that March}.
\end{examples}

Given an anaphor $a$ and its antecedent candidate list $E_a$, we predict the most semantically related NP among all NPs in $E_a$ as the antecedent for $a$. In
case of a tie,  the closest one is chosen to be the predicted antecedent.

The relatedness is measured via cosine similarity between the vector representation of the anaphor and the vector representation of the candidate. 
More specifically, given a noun phrase $np_1$, we first construct a list $N$ which consists of the head and all common nouns (e.g., \emph{\underline{earthquake} victims}), 
adjectives (e.g., \emph{\underline{economical} sanctions}), and ed/ing participles (e.g., \emph{the \underline{collapsed} roadway} and \emph{the \underline{landing} site}) appearing before the head. 
If $np_1$ contains a post-modifier NP $np_2$ via the preposition ``of'', 
we also add the above premodifiers and the head of $np_2$ to the list $N$ (e.g., \emph{the policies of \underline{racial segregation}}).   
Finally, the noun phrase $np_1$ is represented  as a vector $v$ using the following formula, where the suffix ``\_PP'' is added to each $n$ if $np_1$ is a bridging anaphor
and its semantic type is not time:

\begin{equation} \label{eq:vector}
v = \frac{\sum_{n \in N \;} embeddings\_{bridging}_n}{|N|}
\end{equation}

The underlying intuition of adding NP modifications to the list $N$ is that the above mentioned modifiers also represent core semantics of an NP, therefore we should consider them when selecting  
antecedents for bridging anaphors. For instance, as shown in Table \ref{tab:similarity}, for Example \ref{ex:bridging2}, the cosine similarity between \{\textbf{arrangements\_PP}\} and \{\emph{products}\} is 0.06, 
while the cosine similarity between \{\textbf{distribution\_PP}, \textbf{arrangements\_PP}\} and \{\emph{dialysis}, \emph{products}\} is 0.17.

If none of the words in $N$ is present in \emph{embeddings\_bridging}, we simply neglect the noun phrase $np1$.
Note that we do not add the suffix ``\_PP'' to a bridging anaphor representing time information, because such an anaphor is likely to have the same semantic type antecedent (see Example \ref{ex:bridging3}).
Therefore we use semantic similarity instead of relatedness to find its antecedent.

\section{Experiments}
\subsection{Dataset}
For the task of bridging anaphora resolution, we use the dataset ISNotes\footnote{\url{http://www.h-its.org/en/research/nlp/isnotes-corpus}}
  released by \newcite{markert12}.
This dataset contains around 11,000 NPs annotated for information status including 663 bridging NPs
and their antecedents in 50 texts taken from the WSJ portion of the
OntoNotes corpus \cite{ontonotes4.0data}.
As stated in Section \ref{sec:relatedwork},
bridging
anaphors in ISNotes are not limited to definite NPs as in previous work \cite{poesio97b,poesio04d,lassalle11}. 
The semantic relations between anaphor and antecedent 
in the corpus are quite diverse:
only 14\% of anaphors have a part-of/attribute-of relation with the antecedent and only 7\% of anaphors stand in a set relationship to the antecedent. 
79\% of anaphors have ``other'' relation with their antecedents. This includes
encyclopedic or frame relations such as \emph{restaurant}\ -- \textbf{the
  waiter} as well as context-specific relations such as
\emph{palms}\ -- \textbf{the thieves}. In Example \ref{ex:bridging2}, ``\emph{dialysis products}'' is the ``theme'' of the \textbf{distribution arrangements}. 
More specifically, ``\emph{dialysis products}'' belongs to the frame element ``Individuals'' in the ``Dispersal'' frame that is triggered by ``\textbf{distribution arrangements}''. 

\subsection{Experimental Setup}
Following \newcite{houyufang13a}'s experimental setup, we resolve bridging anaphors to
entity antecedents. Entity information is based on the
OntoNotes coreference annotation.
We also use the OntoNotes named entity annotation to assign NPs the semantic type ``\emph{time}'' if their entity types are ``date'' or ``time''.

In \newcite{houyufang13a}, features
are extracted by using entity information. For instance, the raw hit counts of the preposition pattern query (e.g., \textbf{arrangements} of \emph{products})
for a bridging anaphor $a$ and its antecedent candidate $e$ is the maximum count among all
instantiations of $e$. In our experiments, we simply extend the list of antecedent candidates $E_a$ (described in Section \ref{sec:algorithm}) 
to include all instantiations of the original entities in $E_a$. Note that our simple antecedent candidate selection strategy (described in Section \ref{sec:algorithm}) allows us 
to include 76\% of NP antecedents compared to 77\% in \emph{pairwise model III} from \newcite{houyufang13a} where they add top 10\% salient entities as additional antecedent candidates.
In \newcite{houyufang13a}, salient entities on each text are measured through the lengths of the coreference chains based on the gold coreference annotation.

Following \newcite{houyufang13a}, we measure \emph{accuracy}\ on   the number of bridging 
anaphors, instead of on  all links between bridging
anaphors and their antecedent instantiations. We calculate how many
bridging anaphors are correctly resolved among all bridging
anaphors.

\subsection{Using NP Head Alone}
\label{subsec:eval1}
Given an anaphor $a$ and its antecedent candidate list $E_a$, we predict the most related NP among all NPs in $E_a$ as the antecedent for $a$\footnote{In case of a tie, the closest one is chosen to be the
predicted antecedent.}.
The relatedness is measured via cosine similarity between the head of the anaphor (plus the postfix ``\_PP'' if the anaphor is not a time expression) and the head of the candidate.
We run experiments on the following four word embeddings: the original GloVe vectors trained on Gigaword and Wikipedia 2014 dump (\emph{GloVe\_GigaWiki14}), 
GloVe vectors that we trained on Gigaword only (\emph{GloVe\_Giga}), word vectors from \newcite{houyufang18} (\emph{embeddings\_PP}), and our word representation resource 
described in Section \ref{subsec:embeddingsbridging} (\emph{embeddings\_bridging}). Note that for the first two word vectors, we do not add 
the suffix ``\_PP'' to the anaphor's head since such words do not exist in \emph{GloVe\_GigaWiki14} and \emph{GloVe\_Giga}.

Table \ref{tab:result1} lists the results for bridging anaphora resolution based on different word representation resources\footnote{Note that the results for the first three word embeddings are slight better than the ones
reported in \newcite{houyufang18}. This is due to the improved antecedent candidate selection strategy described in Section \ref{sec:algorithm}.}.  
We notice that there is not much difference between \emph{GloVe\_GigaWiki14} and \emph{GloVe\_Giga}.
We find that using \emph{embeddings\_PP} achieves an accuracy of 33.03\% on the ISNotes corpus, which outperforms the results based on \emph{GloVe\_GigaWiki14} and \emph{GloVe\_Giga} by a large margin. 
Using \emph{embeddings\_bridging} further improves the result by 1.8\%. Although the improvement is not significant, we suspect that the representations for words 
without the suffix ``\_PP'' in \emph{embeddings\_bridging} are more accurate because they are trained on many more instances in the vanilla GloVe vectors (\emph{GloVe\_GigaWiki14}).

\begin{table}[t]
\begin{center}
\begin{tabular}{l|l}
\hline
& \textbf{acc}\\ 
\hline
\emph{GloVe\_GigaWiki14}&21.42\\
\emph{GloVe\_Giga}&21.87\\
\emph{embeddings\_PP}&$\textbf{33.03}$\\
\emph{embeddings\_bridging}&$\textbf{34.84}$\\
\hline
\end{tabular}
\end{center}
\caption{\label{tab:result1} Results of using NP head alone for bridging anaphora resolution based on different word representation resources. Bold indicates
statistically significant differences over the baselines (two-sided paired approximate randomization test, $p < 0.01$).}
\end{table}

\subsection{Using NP Head + Modifiers}
\label{subsec:eval2}

We carried out experiments using the deterministic algorithm described in Section \ref{sec:algorithm} together with different word embeddings. 
Again we do not add the suffix ``\_PP'' to the bridging anaphors for \emph{GloVe\_GigaWiki14} and \emph{GloVe\_Giga}.

Table \ref{tab:result2} lists the best results of the two models for bridging anaphora resolution from \newcite{houyufang13a}. \emph{pairwise model III} is a pairwise 
mention-entity model based on various semantic, syntactic and lexical features. \emph{MLN model II} is a joint inference framework based on Markov logic networks \cite{domingos09}. 
It models that semantically or syntactically related anaphors are likely to share the same antecedent and achieves an accuracy of 41.32\% on the ISNotes corpus.

\begin{table}[t]
\begin{center}
\begin{tabular}{l|l}
\hline
& \textbf{acc}\\ 
\hline
\multicolumn{2}{c}{models from \newcite{houyufang13a}}\\ \hline
\emph{pairwise model III}&36.35\\
\emph{MLN model II}&\textbf{41.32}\\
\hline
\multicolumn{2}{c}{NP head + modifiers}\\ \hline
\emph{GloVe\_GigaWiki14}&20.52\\
\emph{GloVe\_Giga}&20.81\\
\emph{embeddings\_PP} &31.67\\
\emph{embeddings\_bridging}&$\textbf{39.52}$\\
\hline
\end{tabular}
\end{center}
\caption{\label{tab:result2} Results of using NP head plus modifications in different word representations for bridging anaphora resolution compared to
the best results of two models from \newcite{houyufang13a}. Bold indicates
statistically significant differences over the other models (two-sided paired approximate randomization test, $p < 0.01$).}
\end{table}

The results for \emph{GloVe\_GigaWiki14} and \emph{GloVe\_Giga} are similar on two settings (\emph{using NP head} vs. \emph{using NP head + modifiers}).
For \emph{embeddings\_PP}, the result on \emph{using NP head + modifiers} (31.67\%) is worse than the result on \emph{using NP head} (33.03\%). 
However, if we apply \emph{embeddings\_PP} to a bridging anaphor's head and modifiers, and only apply \emph{embeddings\_PP} to the head noun of an antecedent candidate, 
we get an accuracy of 34.53\%.
Although the differences 
are not significant, it confirms that the information from the modifiers of the antecedent candidates in \emph{embeddings\_PP} hurts the performance.
This corresponds to our observations in the previous section that the representations  
for words without the suffix ``\_PP'' in \emph{embeddings\_PP} are not as good as in \emph{embeddings\_bridging} due to less training instances. 

Finally, our method based on \emph{embeddings\_bridging} achieves an accuracy of 39.52\%, which is competitive to the best result (41.32\%) reported 
in \newcite{houyufang13a}. There is no significant difference between \emph{NP head + modifiers} based on \emph{embeddings\_bridging} and \emph{MLN model II} (randomization test with $p < 0.01$).

To gain an insight into the contribution of \emph{embeddings\_bridging} on different relation types,
we analyze the results of our method using \emph{embeddings\_bridging} on three relation types: \emph{set-of}, \emph{part-of}, and \emph{other}. 
The accuracies on these three relation types are 17.78\%, 50.0\%, and 39.16\%, respectively. This suggests that in the future
we should include more context for bridging anaphors that hold the \emph{set-of} relation to their antecedents, because the head nouns of such anaphors
often do not bear any specific meanings (e.g., \textbf{Another}). 

\subsection{Analysis of Modifiers}
To better understand the role of NP modifiers in our method, we carried out experiments on \emph{embeddings\_bridging} using different set of modifiers (see Table \ref{tab:result3}). 
It seems that among all three types of modifiers, compared to using NP head alone, adding noun modifiers has the positive impact (36.65\% on \emph{NP head + noun modifiers}
 vs. 34.84\% on \emph{NP head}). Although adding only adjective modifiers does not have influence on results, combining them with noun modifiers yields some improvement over adding only noun 
 modifiers (38.31\% on \emph{NP head + noun\&adjective modifiers} vs. 36.65\% on \emph{NP head + noun modifiers}). On the other hand, 
 ed/ing participle modifiers only have a small positive impact over \emph{NP head + noun modifiers} when combining with noun modifiers.

\begin{table}[t]
\begin{center}
\begin{tabular}{l|l}
\hline
\emph{embeddings\_bridging}& \textbf{acc}\\ 
\hline
\emph{NP head }&34.84\\
\emph{+ all modifiers}&39.52\\
\hline
\emph{+ noun modifiers}&\textbf{36.65}\\
\emph{+ adjective modifiers}&34.84\\
\emph{+ ed/ing participle modifiers}&34.84\\
\emph{+ noun\&adjective modifiers}&\textbf{38.31}\\
\emph{+ noun\&ed/ing participle modifiers}&\textbf{36.80}\\
\emph{+ adjective\&ed/ing participle modifiers}&34.84\\
\hline
\end{tabular}
\end{center}
\caption{\label{tab:result3} Results of using NP head plus different modifications in \emph{embeddings\_bridging}.}
\end{table}

\begin{table*}[thb]
\begin{center}
\begin{tabular}{l|l|l}
\hline
&\textbf{System} &\textbf{acc}\\ 
\hline
\textbf{Baselines} &\emph{\newcite{schulteimwalde98}}&13.68\\
&\emph{\newcite{poesio04d}}&18.85\\
\hline
\textbf{Models from} &\emph{pairwise model III}&36.35\\
\textbf{\newcite{houyufang13a}}&\emph{MLN model II}&41.32\\
\hline
\textbf{\newcite{houyufang18}}&\emph{MLN model II + embeddings\_PP (NP head + noun pre-modifiers)}&\textbf{45.85}\\
\hline 
\textbf{This work}&\emph{embeddings\_bridging (NP head + modifiers)}&39.52\\
&\emph{MLN model II + embeddings\_bridging (NP head + modifiers)}&\textbf{46.46}\\ \hline
\end{tabular}
\end{center}
\caption{\label{tab:result5} Results of different systems for bridging anaphora resolution in ISNotes. Bold indicates
statistically significant differences over the other models (two-sided paired approximate randomization test, $p < 0.01$).}
\end{table*}

\begin{table*}[th]
\begin{center}
\begin{tabular}{l|l|c|l}
\hline
\textbf{Corpus}&\textbf{Bridging Type} &\textbf{\# of Anaphors}& \textbf{acc}\\ 
\hline
\emph{BASHI}&referential, including comparative anaphora&452&27.43\\
\emph{BASHI}&referential, excluding comparative anaphora&344&29.94\\
\emph{ARRAU (RST Train)}&mostly lexical, some referential&2,325&31.44\\
\emph{ARRAU (RST Test)}&mostly lexical, some referential&639&32.39\\
\hline
\end{tabular}
\end{center}
\caption{\label{tab:result6} Results of resolving bridging anaphors in other corpora. Number of bridging anaphors is reported after filtering out a few problematic cases on each corpus.}
\end{table*}

\subsection{Combining \emph{NP Head + Modifiers} with \emph{MLN II}}
For bridging anaphora resolution, \newcite{houyufang18} integrates a much simpler deterministic approach by combining an NP head with its noun modifiers (appearing before the head) based on \emph{embeddings\_PP} into the \emph{MLN II} system \cite{houyufang13a}.
Similarly, we add a constraint on top of \emph{MLN II} using our deterministic approach (\emph{NP head + modifiers}) based on \emph{embeddings\_bridging}. 
Table \ref{tab:result5} lists the results of different systems\footnote{We also reimplement the algorithms from  \newcite{schulteimwalde98} and \newcite{poesio04d} as baselines (Table \ref{tab:result5}). \newcite{schulteimwalde98}  
resolved bridging anaphors to the closest antecedent candidate in a high-dimensional space. We use the 2,000 most frequent words (adjectives, common nouns,
proper nouns, and lexical verbs) from Gigaword as the context words. \newcite{poesio04d} applied a pairwise model combining
lexical semantic features and salience features to perform mereological bridging resolution in the GNOME corpus. We use a Naive  Bayes  classifier  with  standard  settings  in
WEKA \cite{witten05a} and apply the best first strategy to select the antecedent for each anaphor.} 
for bridging anaphora resolution in ISNotes. It shows that combining our deterministic approach (\emph{NP Head + modifiers}) 
with \emph{MLN II} slightly improves the result compared to \newcite{houyufang18}.

Although combining \emph{NP Head + modifiers} with \emph{MLN II} achieves significant improvement over \emph{NP Head + modifiers}, we think the latter has its own value. 
Our deterministic algorithm is simpler and more efficient compared to \emph{MLN model II + embeddings\_bridging}, which contains many complicated features and might be hard to migrate to other bridging corpora. 
Moreover, our algorithm is ``unsupervised'' and requires no training when applied to other English bridging corpora.

\subsection{Resolving Bridging Anaphors in Other Corpora}
Recently, two new corpora containing bridging annotation have become available. The BASHI corpus \cite{roesiger18b} contains 459 bridging NPs and their antecedents in 50 World Street Journal articles. 
Similar to 
ISNotes, BASHI includes both definite and indefinite referential bridging anaphors. 
In addition, comparative anaphora is also considered as bridging anaphora in BASHI. 

Another new corpus for bridging is the
second release of the ARRAU corpus, which contains 5,512 bridging pairs in three different
domains \cite{poesio18}. 
However, most bridging links in ARRAU are purely lexical bridging pairs, and only a small subset of the annotated pairs contains truly
anaphoric bridging anaphors \cite{roesiger18a}. Following \newcite{roesiger18a}, we focus on resolving bridging anaphors in the news text domain (RST).

Based on \emph{embeddings\_bridging}, we apply our deterministic algorithm with small adaptations to resolve bridging anaphors to entity antecedents on the BASHI and ARRAU (RST) corpora. 
Specifically, for the BASHI corpus, we do not add NPs from the first sentence to the list of antecedent candidates $E_a$. 
This is because the phenomenon of globally salient antecedents being linked to all anaphors in a text is less obvious in BASHI. 
In addition, comparative anaphors often have the same semantic class as their antecedents, therefore we do not add the suffix ``\_PP'' to a bridging anaphor if it is a comparative anaphor.

For the ARRAU corpus, we construct the list of antecedent candidates $E_a$ using NPs preceding $a$ from the same sentence as well as from the previous ten sentences.
Since most bridging pairs in ARRAU are lexical bridging (e.g., \emph{Tokyo} -- \textbf{Japan}, \emph{other nations} -- \textbf{Britain}) and anaphors often 
have the same semantic type as their antecedents, we do not add the suffix ``\_PP'' to bridging anaphors.

Table \ref{tab:result6} lists the results of bridging anaphora resolution in the BASHI and ARRAU corpora, respectively. 
On the test set of the ARRAU (RST) corpus, \newcite{roesiger18c} proposed a modified rule-based system based on \newcite{houyufang14}'s work and reported 
an accuracy of 39.8\% for bridging anaphora resolution. And our algorithm achieves an accuracy of 32.39\% using only \emph{embeddings\_bridging}. 
Overall, the reasonable performance on these two corpora demonstrates that \emph{embeddings\_bridging} is a general word representation resource for bridging.


\section{Conclusions}
We improve the word representation resource \emph{embeddings\_PP} \cite{houyufang18} by combining it with \emph{GloVe}. 
The resulting word embeddings (\emph{embeddings\_bridging}) are a more general word representation resource for bridging.
Based on \emph{embeddings\_bridging}, we propose a  
deterministic approach for choosing antecedents for bridging anaphors. We show that this simple and efficient method achieves the 
competitive result on bridging anaphora resolution compared to the advanced machine learning-based approach in \newcite{houyufang13a} which is heavily dependent on 
a lot of carefully designed complex features. 
We also demonstrate that using \emph{embeddings\_bridging} yields better results than using \emph{embeddings\_PP} for bridging anaphora resolution.


For the task of bridging anaphora resolution, \newcite{houyufang13a} pointed out that considering only head noun knowledge is not enough and future work needs to explore wider context to resolve context-specific bridging relations.
In this work we explore the context within NPs---that is,  we combine the semantics of certain modifications and the head by vector average using \emph{embeddings\_bridging}. 
But in some cases, knowledge about NPs themselves is not enough for resolving bridging. For instance, in Example \ref{ex:bridging4}, 
knowing that \textbf{any loosening} has the ability to ``rekindle inflation'' from 
the context of the second sentence 
can help us to find its antecedent ``\emph{the high rates}'' (which is used to against inflation).

\begin{examples}
\item \label{ex:bridging4} Chancellor of the Exchequer Nigel Lawson views \emph{the high rates} as his chief weapon against inflation, which was ignited by tax cuts and loose credit policies in 1986 and 1987. 
Officials fear that \textbf{any loosening} this year could rekindle inflation or further weaken the pound against other major currencies.
\end{examples}

In the future, we will study how to integrate context outside of NPs for the task of choosing antencedents for bridging anaphors. 
Also we hope that our word representation resource will facilitate other related research problems such as semantic role labeling.






\section*{Acknowledgments}
The author appreciates the valuable feedback from the anonymous reviewers
and would like to thank Massimo Poesio for sharing the ARRAU corpus.

\bibliographystyle{acl_natbib_nourl}
\bibliography{../../bib/lit/lit}
\end{document}